\documentclass[letterpaper, 10 pt, conference]{ieeeconf}  

\IEEEoverridecommandlockouts                              

\usepackage{balance}
\usepackage{lmodern}
\usepackage{graphicx}
\usepackage{amsmath}
\usepackage{amssymb} 
\usepackage{booktabs}
\usepackage{tcolorbox}
\usepackage{hyperref}
\usepackage{tabularx}
\usepackage{subcaption}
\usepackage{xcolor}
\usepackage{eso-pic}

\usepackage{mathptmx} 

\definecolor{myblue}{RGB}{0, 0, 255} 
\definecolor{mypink}{RGB}{255, 182, 193} 
\definecolor{mygreen}{RGB}{27, 120, 55}
\definecolor{myorange}{RGB}{230, 97, 1}

\title{\LARGE \bf
When and How to Express Empathy in Human-Robot Interaction Scenarios
}

\author{
Christian Arzate Cruz$^{1}$, 
Edwin C. Montiel-Vazquez$^{2}$, 
Chikara Maeda$^{1}$, 
and Randy Gomez$^{1}$%
\thanks{$^{1}$Honda Research Institute Japan (HRI-JP), Wako City, Japan.}
\thanks{$^{2}$Tecnológico de Monterrey, Monterrey, Mexico.}
\thanks{\tt\small christian.arzate@jp.honda-ri.com}
}

\newcommand\copyrightnotice{%
  \AddToShipoutPictureFG*{%
    \AtPageLowerLeft{%
      \parbox[b][2cm][c]{0.95\paperwidth}{\centering
        \footnotesize \textcopyright~2025 IEEE. 
        Personal use of this material is permitted. The final version was published in the 
        Proceedings of the IEEE nternational Symposium on Robot and Human Interactive Communication (RO-MAN) 2025. 
        Reference: Christian Arzate Cruz et al., “When and How to Express Empathy in Human-Robot Interaction Scenarios,” \textit{RO-MAN 2025}. IEEE Press.}%
    }%
  }%
}

\begin{document}
\maketitle
\copyrightnotice

\begin{abstract}
Incorporating empathetic behavior into robots can improve their social effectiveness and interaction quality. In this paper, we present \textit{whEE} (\textbf{w}hen and \textbf{h}ow to \textbf{e}xpress \textbf{e}mpathy), a framework that enables social robots to detect when empathy is needed and generate appropriate responses. Using large language models, \textit{whEE} identifies key behavioral empathy cues in human interactions. We evaluate it in human-robot interaction scenarios with our social robot, Haru. Results show that \textit{whEE} effectively identifies and responds to empathy cues, providing valuable insights for designing social robots capable of adaptively modulating their empathy levels across various interaction contexts.
\end{abstract}


\section{Introduction} 

In most scenarios, Large Language Models (LLMs) represent the state-of-the-art approach for classifying empathy~\cite{Shen2024empathicstories, Barros2019omg} and generating empathetic responses~\cite{Zhang2024stickerconv, Ma2020survey}. However, the development of robots capable of dynamically adjusting their level of empathy based on the context remains an underexplored area~\cite{Arzate2025_hri}. To this end, we introduce \textit{whEE} (\textbf{w}hen and \textbf{h}ow to \textbf{e}xpress \textbf{e}mpathy), an empathy framework that provides guidelines on \textit{when} robots should respond empathetically and \textit{how} to achieve it. Using our framework, we analyze the utterances of speakers and listeners in dyadic and group conversations with varying levels of empathy. Our analysis identifies key empathy cues that indicate when a speaker seeks an empathetic response and the cues exhibited by listeners displaying high levels of empathy.

We approach empathy by focusing on observable behaviors that individuals exhibit when demonstrating an understanding of others' emotions and engaging deeply with their experiences—referred to as behavioral empathy~\cite{clark_i_2019}.

Previous research has investigated the detection of empathy direction—whether a person is seeking or providing empathy—using datasets such as iEmpathize~\cite{Hosseini2021takes} and TwittEmp~\cite{hosseini-caragea-2021-distilling-knowledge}. However, these studies have not included interactions from human-robot interaction (HRI) contexts and have yet to explore the effectiveness of LLMs in identifying empathy direction. This gap motivates our study, in which we apply our framework, \textit{whEE}, specifically to HRI scenarios involving dyadic and group human-human interactions mediated by our social robot, Haru~\cite{Cooper2024}. We use Haru as it aims to foster safe environments for improving human-human connection.   


Additionally, \textit{whEE} includes an empathetic text-generation module designed to enhance robot empathy. Using prerecorded human interactions as input, we systematically prompt empathetic responses from LLMs for Haru. We then compare these empathetic responses to Haru's standard responses to evaluate their differences.

With the development of our empathy framework for HRI, \textit{whEE}, we contribute to the literature by offering new insights into empathetic human-human and human-robot interactions. Besides, our findings can support researchers in designing robots that provide better empathy only when needed.

\section{Related Work} 


Detection~\cite{Hasan2023empathy} and replication of empathetic behavior~\cite{raamkumar_empathetic_2023} have been studied in computing using text. However, these tasks are rarely explored in HRI settings.

\subsection{Empathy}


Empathy has many different definitions that vary according to their setting and application~\cite{Cuff2016,clark_i_2019}. One method to incorporate empathy into robotics and computing is to adopt a psychology-based empathy definition~\cite{Hosseini2021takes,Spring2019,montiel-vazquez_explainable_2022}. Meanwhile, other works attempt to propose definitions of empathy specific to their computing application~\cite{ab_aziz_conceptual_2022,paiva_empathy_2017, brannstrom_formal_2024,yalcin_modeling_2020}. In our work, we use a definition presented in~\cite{Montiel2024} and merge it with the work by Hosseini et al.~\cite{hosseini_it_2021,hosseini-caragea-2021-distilling-knowledge}. We present our formal empathy definition in Section \ref{sec:background}.


\subsubsection{Empathy Cues}

Previous work in computing has attempted to identify the mechanisms, concepts, and components that characterize empathetic responses. Among these, the work of Raamkumar et al.~\cite{raamkumar_empathetic_2023} has explored the most popular cues used to generate empathetic text, with emotion being one of the most widely adopted cues. Likewise, research focused on empathy detection~\cite{Montiel2024,hosseini-caragea-2021-distilling-knowledge} has identified that sentiment and emotion are very relevant variables when identifying responses that provide empathy. Finally, the work by Sharma et al.~\cite{sharma-etal-2020-computational} presents one of the most popular methods to describe empathy in terms of specific communication mechanisms: The EPITOME framework.

\subsection{Empathetic Robots}
The HRI community has explored the use of empathetic robots that maintain consistent behavior over time~\cite{Gonsior2012,Cramer2010effects}. For example, authors in~\cite{Alves2019empathic} designed a multiplayer game mediated by a robot, which performs its role using handcrafted empathetic behaviors based on demonstrations from human experts.

The authors of~\cite{Arzate2025_hri} proposed a robot with an adaptive empathy level designed for dyadic conversations between a robot and a human. At each conversational exchange, the human's empathy level is computed and integrated into the prompt used to generate the robot's empathetic behavior, enabling the robot to dynamically match the human's empathy level. In contrast, our paper employs a different strategy to identify moments when individuals seek empathy. We utilize human-annotated datasets explicitly labeled with situations in which a person anticipates receiving an empathetic response. Additionally, we analyze the observable empathy cues exhibited when a person seeks or provides empathy.

\subsection{Empathy Classifiers}
Empathy classification has been attempted in various contexts. Hasan et al.~\cite{Hasan2023empathy} identified that most empathy classification approaches use BERT, RoBERTa, and its derivatives for empathy prediction. The scenarios for empathy classifiers have ranged from HRI~\cite{Barros2019omg} to Healthcare~\cite{sharma-etal-2020-computational,hosseini_it_2021, lee2024acnempathize}. Most empathy classification focuses on the detection of empathy by the one who is providing it, either internally felt~\cite{Shen2024empathicstories}  or externally expressed~\cite{montiel-vazquez_explainable_2022,Montiel2024}. However, there are very few classifiers for seeking empathy. For this task, the work by Hosseini and Caragea~\cite{hosseini_it_2021,hosseini-caragea-2021-distilling-knowledge} presents the most prominent approaches: a binary label that identifies utterances in which someone seeks an empathetic response. This was relevant for our \textit{why}, as this task is yet to be explored in the context of HRI. 

LLMs for empathy classification seem to have potential~\cite{hasan-etal-2024-llm,Hasan2023empathy,xie_scoring_2024}. Therefore, we are interested in exploring their feasibility in HRI. 

\subsection{Empathetic Text Generation}
Previous surveys on empathetic conversational systems (ECS) have identified that including external information, such as emotion causes or cognitive knowledge, improves the quality of responses~\cite{raamkumar_empathetic_2023,guo2024review}. However, one of the gaps identified in this research continues to be the integration of ECS into specific applications. Most approaches for empathetic text generation are trained and evaluated on the work of Rashkin et al.~\cite{rashkin_towards_2019}, with the EmpatheticDialogues (ED) dataset and the Transformer~\cite{Vaswani2017} family of architectures. 

The literature indicates that LLMs can produce empathetic responses that surpass human capabilities in specific contexts, such as healthcare and legal fields, without compromising factual accuracy~\cite{sorin_large_2024, lee_large_2024, zhang_empathetic_2024}. Nevertheless, multiple strategies have been proposed to enhance their performance, including the integration of external knowledge~\cite{qian_harnessing_2023} and the use of alternative communication channels like stickers~\cite{zhang_stickerconv_2024}. Previous studies confirm the capacity of LLMs to deliver empathetic responses. However, Cuadra et al.~\cite{Cuadra2024illusion} highlighted a significant limitation: LLMs may provide empathy indiscriminately, even toward identities or viewpoints associated with harmful ideologies. Consequently, there is a clear need to further explore and expand research on finding the right moments to respond empathetically.

\section{The whEE Framework} 
\begin{figure}[tbh]
    \centering
    \includegraphics[width=0.95\linewidth]{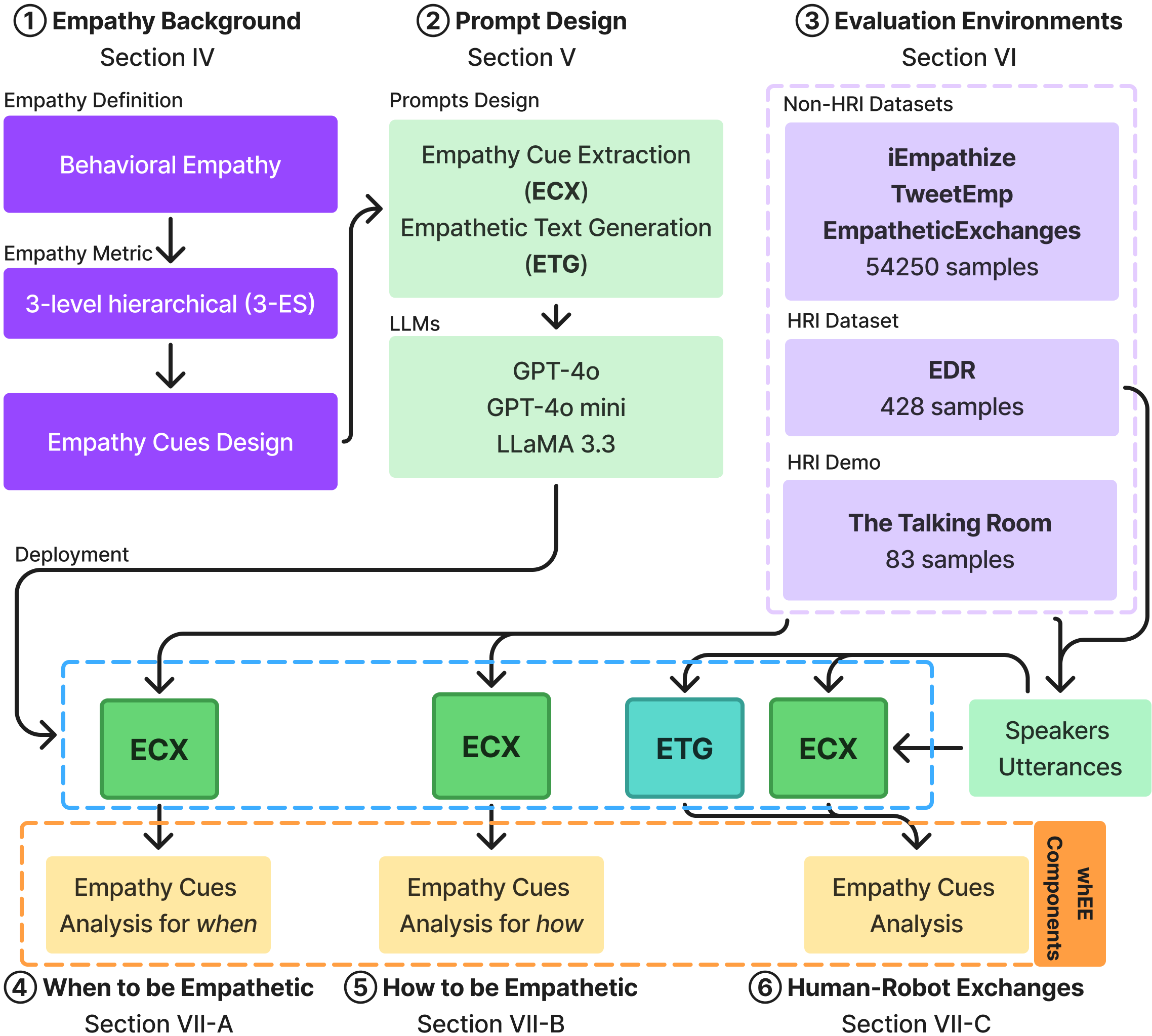}
    \caption{Overview of the design of our \textit{whEE} framework.}
    \label{fig:how}
\end{figure}

In this section, we provide an overview of the design of our three component empathy framework for HRI, \textit{whEE}. Figure~\ref{fig:how} illustrates the key components of our framework and its six-step development process. We intend for \textit{whEE} to serve as a tool that helps HRI researchers design empathy-centric applications and explore new research avenues on how empathy in HRI impacts the quality of interactions.

The \textit{whEE} framework provides insight into the observed empathy cues when someone initiating a conversation -- the speaker -- hopes to receive an empathetic response from their conversation partner -- the listener. The idea is to identify empathy cues from speakers seeking empathy, helping determine appropriate moments for empathetic responses. Additionally, \textit{whEE} captures cues from listeners providing empathy. These two components represent the \textit{when} and \textit{how} of expressing empathy.

The first two components of \textit{whEE} (steps $4$ and $5$) focus on analyzing empathy cues at different stages of a conversation and at various empathy levels. The third component, human-robot exchanges, aims to study how LLMs generate empathetic text in human-robot interaction scenarios (step $6$). The goal of this last component is to describe the empathy cues exhibited by LLMs when they are prompted to be empathetic toward humans. For this analysis, we test different prompts (step $2$) designed based on our definition of empathy cues (step $1$).

\subsection{When}
In the context of our research, by \textit{when}, we refer to exchanges in which the \textit{speaker presents an opportunity to receive empathy}. This means we are interested in empathy cue exchanges that aim to mimic those observed in everyday human-human conversations.

Although we could design robots that consistently exhibit a high level of empathy, constant exposure to empathetic exchanges can make humans susceptible to exhaustion or empathy burnout~\cite{Thomas2007distress, Omdahl1999emotional, Nourriz2024design}. For this reason, empathy modulation is key to developing robots that can identify the appropriate moments to empathize with humans.

Various factors can influence when humans empathize, such as personality~\cite{Eisenberg2001origins} and past experiences~\cite{Nichols1996varieties}. \textit{whEE} focuses on empathy modulation based on intensity and empathy cues displayed by the speaker.

\subsubsection{Assumptions}
First, the \textit{whEE} framework assumes that the robot's primary role is to foster and enhance its relationship with humans over time, making empathy a crucial component of its programming. Additionally, the framework assumes that the robot's empathetic capabilities remain consistent throughout interactions. However, integrating an empathy modulation module that adapts based on the evolving affective bond between humans and robots presents a promising research direction in HRI, which researchers can further explore utilizing our proposed framework.

\subsection{How}
In our paper, the \textit{how} refers to \textit{the nature of empathetic responses from listeners}. Our goal is to identify patterns within textual responses that characterize empathy. To achieve this, we examine responses for the presence or absence of specific empathy cues, such as exploration. We formally define our set of empathy cues in Section~\ref{sec:empathy_cues}.

In affective computing, it is common to find approaches that evaluate empathy by considering both the speaker's and the listener's utterances. However, our approach focuses exclusively on the listener's utterance. Specifically, we adopt a strategy similar to the one proposed in~\cite{Hosseini2021takes,hosseini-caragea-2021-distilling-knowledge}, enabling us to use their datasets in that format.

Not all empathetic responses exhibit the same empathy cues, as individuals can respond at varying empathy levels. For instance, according to a metric based on the three-level Russian doll model~\cite{waal_russian_2007,Montiel2024}, level $1$ (little to no empathy) involves emotional contagion, level $2$ (somewhat empathetic) includes responses showing interest and consolation, and level $3$ (highly empathetic) encompasses genuine perspective-taking and an intention to assist the speaker. The \textit{whEE} binary labels classify responses as either level $1$ for non-empathetic or level $2$ for empathetic. These binary labels correspond respectively to level $1$ (non-empathetic) and levels $2$ and $3$ combined (empathetic) from the original three-level Russian doll model.

\subsubsection{Assumptions}
The \textit{whEE} framework assumes that our selection of the most relevant empathy cues is sufficient to model empathetic responses effectively in both non-HRI and HRI settings. See Section \ref{sec:empathy_cues} for our empathy cues definition.

\subsection{Human-Robot Exchanges}
The \textit{whEE} framework includes an empathetic text generation module designed for our tabletop robot, Haru. Our objective is to develop a text generation module capable of adjusting its empathy level based on the speaker's needs. Specifically, we aim for Haru to demonstrate empathy only when an individual explicitly seeks empathetic responses.

The prompt for Haru will incorporate the most common empathy cues identified in our analysis of \textit{how} to be empathetic. In situations where empathy is not required, Haru will use the standard prompt previously applied in our demos and experiments~\cite{Cooper2024}.

\subsubsection{Assumptions}
To evaluate our empathetic text generation prompt, we will use speaker utterances from our HRI datasets (see Section~\ref{sec:datasets}) as input, effectively replacing human listeners with Haru. Therefore, for this experiment, the \textit{whEE} framework assumes that humans display similar empathy cues when interacting with robots. Further experiments will be necessary in the future to assess how our adaptive text generation prompt performs in different scenarios.

\section{Empathy Background} 
\label{sec:background}

Empathy is a hierarchical construct with three dimensions: feeling like someone else (affective empathy), understanding the affective state of another (cognitive empathy), and behavioral empathy~\cite{clark_i_2019,Cuff2016}. Artificial agents are unable to truly feel or understand affective states. Therefore, we base \textit{whEE} around behavioral empathy, which refers to providing communication and behaviors consistent with the affective and cognitive dimensions, mimicking them. This allows us to conceptualize artificial empathy in a manner consistent with previous research~\cite{cliffordson_hierarchical_2002, waal_russian_2007}.

\subsection{Empathy Cues}
\label{sec:empathy_cues}
In our paper, we define empathy cues as empathy-related concepts used to characterize and classify empathy directions. Various empathy cues have been utilized in previous works, such as intent classification~\cite{welivita_taxonomy_2020} and sentiment analysis~\cite{Montiel2024}. In this study, we employ a set of six empathy cues.

We begin by defining our empathy cues to specially identify when a person is \textbf{seeking empathy}, which are based on the approach proposed by~\cite{Hosseini2021takes}. First, (1) we use a vector model of emotion to capture how empathy directions correlate to the affective dimension of empathy. This model uses \textit{arousal} and \textit{valence} as vectors. Both arousal and valence have values ranging from $-1$ to $1$, representing the intensity of emotion and its positivity or negativity, respectively. Second, (2) we include \textit{sentiment polarity} to identify the overall sentiment expressed, which can be positive, negative, or neutral. Lastly, (3) we identify the \textit{main subject} of an utterance to clarify the primary focus, with possible values being $0$ when the speaker's attention is primarily on themselves, $1$ when the attention is on the conversation partner, and $2$ when it concerns another person or topic.

For concepts related to \textbf{providing empathy}, we use the EPITOME mechanisms proposed in~\cite{sharma-etal-2020-computational}: (4) emotional reactions, (5) interpretations, and (6) explorations. Mechanism~(4) measures the emotional expressiveness of listeners. Mechanism~(5) evaluates how well listeners understand the speaker's emotions and situations. Mechanism~(6) assesses how effectively listeners explore the speaker's emotions. Each mechanism has three possible values: $0$, $1$, and $2$. Level $0$ indicates the absence of the mechanism (e.g., an utterance without any emotional reaction), while levels $1$ and $2$ represent \textit{weak} and \textit{strong} reactions, respectively. A weak emotional reaction does not explicitly mention emotions, whereas a strong emotional reaction explicitly references emotions (e.g., ``I feel sad for you'').

\section{Prompt Design} 
\label{sec:prompt_design}
In this section, we describe our reasoning behind the design choices made for the prompts used to classify the direction of empathy (seek or provide), identify empathy cues, and generate adaptive empathetic text. All our algorithms, models, empathy cue definitions, complete prompts, and datasets are found in our public repository\footnote{\href{https://github.com/christianharu/WHEE}{https://github.com/christianharu/WHEE}}.

\subsection{Classification}
We begin by explaining the main task for the LLM, which involves classifying the direction of empathy into one of three possible labels: \textit{seeking}, \textit{providing}, or \textit{none}. Next, we instruct the LLM to consider our defined set of empathy cues when performing this classification. Each empathy cue includes a definition and a set of possible values. Finally, we provide the utterance to be classified and ask the LLM to generate its classification using a structured format, which includes the values for all empathy cues along with the empathy direction label.

\subsection{Text Generation}

To produce empathetic text, we designed a prompt that integrates the EPITOME mechanisms~\cite{sharma-etal-2020-computational}, each defined at three levels: \textit{none}, \textit{weak}, and \textit{strong}. Using the result of the classification prompt and the speaker's utterance as input, we instruct the LLM to generate empathetic text only when the speaker is seeking empathy; otherwise, a standard non-empathetic prompt is applied. For empathetic responses, we configure the prompt to set all EPITOME mechanisms at the \textit{strong} level.

\section{Evaluation Environments} 
We evaluate the \textit{whEE} model in both non-HRI and HRI scenarios. The non-HRI scenarios include datasets comprising data collected from social media, while the HRI scenarios involve interactions between humans in realistic human-robot interaction contexts. We describe both evaluation environments in detail in the remainder of this section.

\subsection{Non-HRI Datasets}
\label{sec:datasets}

To determine when to offer empathy, we explored the work of Hosseini and Caragea~\cite{Hosseini2021takes,hosseini-caragea-2021-distilling-knowledge}. Their databases, \textit{iEmpathize} and \textit{TweetEmp}, provide labels that mark an utterance as \textit{seeking empathy}, and \textit{providing empathy}, alongside data points that do neither. These databases were obtained from the \textit{Cancer Survivor Network (CSN)} and the social media site \textit{Twitter}. The context of utterances is conversations between cancer patients and survivors.

Despite its labels, both \textit{iEmpathize} and \textit{TweetEmp}~\cite{Hosseini2021takes,hosseini-caragea-2021-distilling-knowledge} are limited since they deal predominately with empathy with a negative sentiment. Likewise, they are focused on cancer survivors. As such, we compliment these datasets with more versatile conversations for social settings that include positive empathy ~\cite{morelli_emerging_2015}. We do this by using \textit{EmpatheticExchanges (EX)}~\cite{Montiel2024}. 

EX consists of exchanges marked with 3-level labels based on the expression of empathy in each exchange: little to no empathy (1), somewhat empathetic (2). and empathetic (3). This data is based on the ED dataset \cite{rashkin_towards_2019}. Therefore, the beginning of each conversation was designed to elicit an empathetic response from another participant using various emotional contexts. We adapt the labels found in EX to match those in \textit{iEmpathize} and \textit{TweetEmp}. Those utterances from the \textit{speaker} role in EX that start a conversation using an emotional prompt are labeled as \textit{seeking} empathy. Meanwhile, those utterances by the \textit{listener} role with the empathy level 2 or 3, are labeled as \textit{providing} empathy. The rest of the utterances were given a label for \textit{none}. Using this labeling scheme, we obtained a combined dataset with $54249$ utterances with rich emotional and contextual diversity and cases of negative and positive empathy. To our knowledge, this is the largest labeled collection of utterances that differentiates seeking empathy as a distinct label. We use this dataset to explore the performance of LLMs on the whEE framework. To do our test, we perform a dataset split of 80/10/10 for training, evaluation, and validation sets. 

\subsection{HRI Datasets}
\subsubsection{Empathetic Dialogues with a Robot}

\begin{figure}[h]
    \centering
    \includegraphics[width=0.65\linewidth]{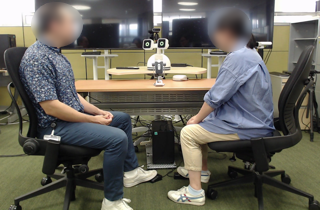}
    \caption{The EDR dataset setting.}
    \label{fig:eerobot}
\end{figure}

The first dataset, \textbf{E}mpathetic \textbf{D}ialogues with a \textbf{R}obot (EDR), comprises semi-structured conversations between a speaker and a listener held in the presence of Haru. The role of Haru was limited to giving an introduction and an emotion for prompting the speaker in each conversation. Figure~\ref{fig:eerobot} illustrates the setup used during data collection.

EDR is based on ED ~\cite{rashkin_towards_2019}. Therefore, we follow the same scheme and create conversation prompts using real-life scenarios corresponding to eight basic emotions ~\cite{Plutchik2001nature}. Participants acting as speakers used these scenarios and their personal experience to start the conversations. Listeners responded at varying empathy levels by role-playing different relationship types—strangers, coworkers, or close friends/family—corresponding to empathy levels $1$, $2$, and $3$, respectively~\cite{waal_russian_2007}. Two experienced annotators assigned empathy labels after receiving training with definitions and examples of empathy levels from the EX dataset~\cite{Montiel2024}. They evaluated listener behavior at the conversation level. The process involved independent annotations, followed by two rounds of discussions to achieve full agreement.

Two male authors served as speakers, aged between $19$–$30$ and $30$–$40$. Twelve listeners (four female), recruited through institutional channels, volunteered for the study. All listeners had previous robot interaction experience and diverse cultural backgrounds. Ethical approval was granted by our institution, and informed consent was obtained from all participants. 

We labeled all speaker utterances as \textit{seeking empathy}. Listener responses with empathy level $1$ were labeled as \textit{none}, while responses at levels $2$ and $3$ were labeled as \textit{providing empathy}. The EDR dataset is available in our repository. It comprises $79$ conversations, $214$ exchanges (2-4 per conversation), and $428$ labeled utterances. Utterances are not longer than $127$ words.

\subsubsection{The Talking Room}
\begin{figure}[h]
    \centering
    \includegraphics[width=0.70\linewidth]{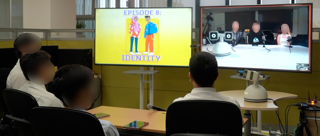}
    \caption{The Talking Room.}
    \label{fig:talking_room}
\end{figure}

The second dataset is derived from our multi-group application, The Talking Room~\cite{Cooper2024}, in which two groups of children engage in $\approx 40$ minute conversation via a video call, each group facilitated by a Haru robot serving as a mediator, as shown in Figure~\ref{fig:talking_room}. Ethical approval was granted by our institution and the schools involved in the study. However, the interaction data involving children cannot be shared due to privacy restrictions.

The Talking Room involved thirteen children from four countries: Japan (3), Australia (4), Namibia (3), and Denmark (3). Participants, aged between $10$ and $15$, provided informed consent individually or through parental approval. Across two initial interaction sessions—Session 1 (Namibia-Denmark, $43$ exchanges) and Session 2 (Japan-Australia, $40$ exchanges)—we collected a total of $164$ utterances. This marked their first experience interacting with peers from the other country.

We manually labeled each utterance to indicate whether the participant was \textit{seeking empathy}, \textit{providing empathy}, or \textit{none}. All the labels are available in our repository.

\section{Evaluation}


The evaluation of \textit{whEE} involves testing the performance of various machine learning-based methods in identifying \textit{when} and \textit{how} to be empathetic. By \textit{when}, we refer to correctly classifying utterances where a person is \textit{seeking empathy} (Label $1$). The \textit{how} focuses on detecting when a person is \textit{providing empathy} (Label $2$). We include utterances where neither is true (Label $0$). Therefore, we report results treating it as a single $3$-label classification task.

We evaluated seven methods covering a range of approaches, including classic artificial intelligence (AI) algorithms, language models (LMs), and LLMs. For classic AI, we used a random forest (RF) classifier and XGBoost~\cite{Chen2016xgboost}. In the LM category, we included BERT-based models~\cite{Devlin2012}: BERT, Modern BERT (MBERT), and RoBERTa. For LLMs, we tested LLaMA~3.3~\cite{meta2024llama3} and two OpenAI models~\cite{openai2024chatgpt} (GPT4o-mini and GPT4o). Performance was measured using the macro-average of precision (p), recall (r), and F1-score (F1). A summary of results for both non-HRI and HRI settings is shown in Table~\ref{tab:results}.

\begin{table}[!tbh]
    \centering
    \caption{When and how to be empathetic classification performance. For each column, the best result is \textbf{bolded}, and the second best is \underline{underlined}.}
    \label{tab:results}
    \begin{tabularx}{\columnwidth}{l *{6}{>{\centering\arraybackslash}X}}
        \toprule
        & \multicolumn{3}{c}{\textbf{Non-HRI}} & \multicolumn{3}{c}{\textbf{HRI}} \\
        \cmidrule(lr){2-4} \cmidrule(lr){5-7}
        & p & r & F1 & p & r & F1 \\
        \midrule
        RF & 0.6496 & 0.6355 & 0.6355 & 0.6002 & 0.4376 & 0.3843\\
        XGBoost & 0.6679 & 0.6498 & 0.6630 & 0.5920 & 0.5000 & 0.4761 \\
        \midrule
        BERT & 0.7571 & 0.7776 & 0.7644 & 0.6586 & 0.5887 & 0.5972 \\
        MBERT & \underline{0.7715} & \textbf{0.8017} & \underline{0.7791} & 0.6392 & 0.5769 & 0.5848 \\
        RoBERTa & \textbf{0.7740} & \underline{0.7998} & \textbf{0.7813} & \underline{0.6701} & 0.5951 & 0.5928 \\
        \midrule
        LLaMA 3.3 & 0.5728 & 0.5841 & 0.5517 & 0.6683 & \textbf{0.6662} & \textbf{0.6595} \\
        GPT4o-mini & 0.5657 & 0.4784 & 0.4742 & 0.6696 & 0.5427 & 0.5407 \\
        GPT4o & 0.5928 & 0.5482 & 0.5490 & \textbf{0.6830} & \underline{0.5964} & \underline{0.6045} \\
        \bottomrule
    \end{tabularx}
\end{table}

\begin{figure*}[!tbh]
    \centering
    \begin{subfigure}[b]{0.49\linewidth}
        \centering
        \includegraphics[width=\linewidth]{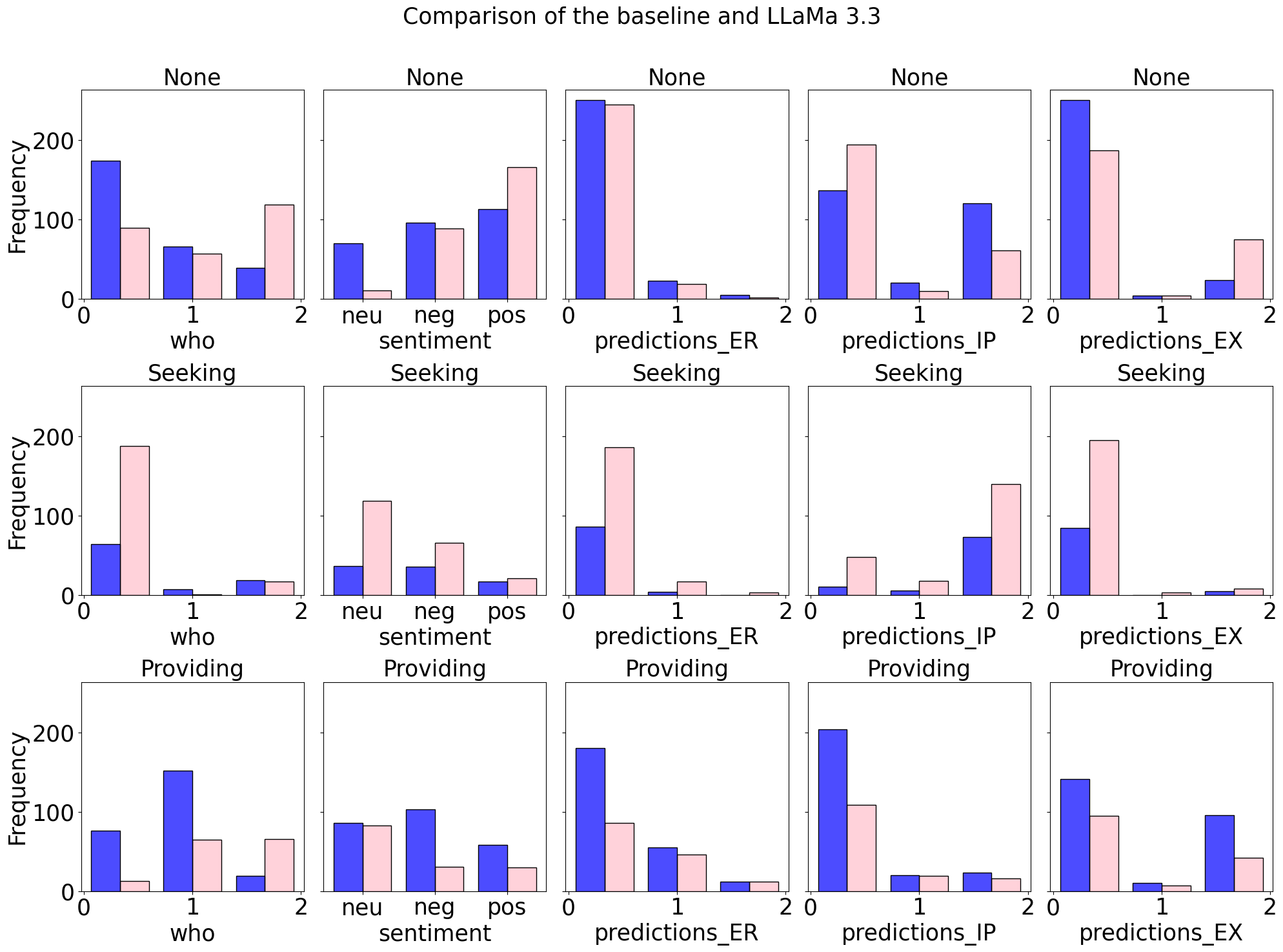}
        \caption{HRI scenarios.}
        \label{fig:cues_hri}
    \end{subfigure}
    \hfill
    \begin{subfigure}[b]{0.49\linewidth}
        \centering
        \includegraphics[width=\linewidth]{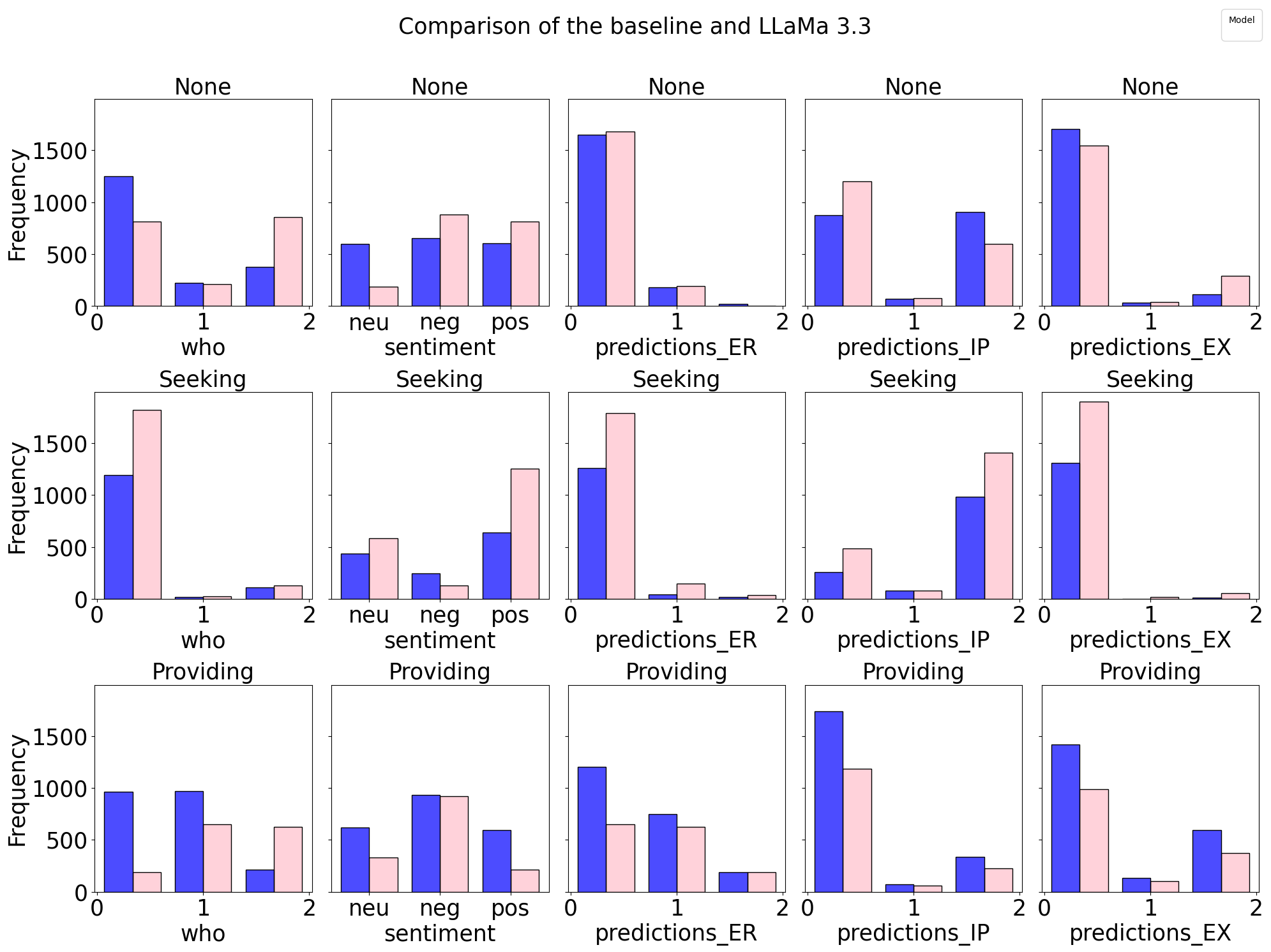}
        \caption{Non-HRI scenarios.}
        \label{fig:cues_non_hri}
    \end{subfigure}
    \caption{Empathy cues distribution comparison across HRI and non-HRI scenarios. Baseline methods (established affective computing methods) are shown in \textcolor{myblue}{\textbf{blue}} and LLaMA~3.3 in \textcolor{mypink}{\textbf{pink}}.}
    \label{fig:cues_comparison}
\end{figure*}

Another key component of \textit{whEE} is the analysis of empathy cues when individuals are seeking or providing empathy. As described in Section~\ref{sec:prompt_design}, part of the LLM prompt instructs the model to generate these cues. For comparison, we also generated them using established affective computing methods, which we refer to as \textit{baseline methods} throughout the rest of the paper. Arousal and valence scores were obtained using the NRC\_VAD Lexicon~\cite{mohammad_obtaining_2018}, while sentiment polarity was derived from a RoBERTa-base model~\cite{Barbieri2020tweeteval}. Emotional reactions, interpretations, and explorations were extracted using EPITOME~\cite{sharma-etal-2020-computational}. In our setup, EPITOME was applied using a dummy speaker utterance (``This is a dummy text''), and we verified that this approach did not affect its performance on the original dataset. Figure~\ref{fig:cues_hri} presents histograms summarizing the distribution of empathy cue values for each class in HRI scenarios, while Figure~\ref{fig:cues_non_hri} shows the corresponding results for non-HRI settings.  We present the frequency distributions for the baseline methods, which rely on standard tools in affective computing, alongside one of best-performing model in our evaluation: LLaMA~3.3. We selected this LLM for comparison due to its strong performance and its use in our robot, Haru.

\subsection{When to be Empathetic} 
Table~\ref{tab:when_empathetic} presents the frequency and proportion (in parentheses) of each empathy cue observed when individuals \textit{seek empathy}, comparing results between non-HRI and HRI scenarios using LLaMA~3.3.

\begin{table}[htb]
\centering
\caption{When to be empathetic cues.}
\label{tab:when_empathetic}
\resizebox{\columnwidth}{!}{%
\begin{tabular}{lcc|cc}
\toprule
\textbf{Empathy Cue} & \textbf{Non-HRI} & \textit{Label} & \textbf{HRI} & \textit{Label} \\
\midrule
Who                 & 1843 (0.92)   & I or We  & 188 (0.91) & I or We \\
Sentiment           & 1269 (0.64)   & Negative & 119 (0.58) & Negative \\
Emotional reaction  & 1810 (0.91)   & $0$ & 186 (0.90) & $0$ \\
Interpretations     & 1418 (0.71)   & Strong & 140 (0.68) & Strong \\
Exploration         & 1919 (0.96)   & $0$ & 195 (0.95) & $0$ \\
\midrule
Valence (mean ± std) & 0.71 ± 0.33 &  & 0.65 ± 0.36 &  \\
Arousal (mean ± std) & 0.52 ± 0.21 &  & 0.50 ± 0.22 &  \\
\bottomrule
\end{tabular}%
}
\end{table}

\subsection{How to be Empathetic}
In contrast, Table~\ref{tab:how_empathetic} presents the empathy cue results for instances where individuals \textit{provide empathy}, based on outputs generated by LLaMA~3.3.

\begin{table}[htb]
\centering
\caption{How to be empathetic cues.}
\label{tab:how_empathetic}
\resizebox{\columnwidth}{!}{%
\begin{tabular}{lcc|cc}
\toprule
\textbf{Empathy Cue} & \textbf{Non-HRI} & \textit{Label} & \textbf{HRI} & \textit{Label} \\
\midrule
Who                 & 671 (0.45)   & You  & 66 (0.46) & Another \\
Sentiment           & 941 (0.62)   & Positive & 83 (0.58) & Positive \\
Emotional reaction  & 668 (0.44)   & $0$ & 86 (0.60) & $0$ \\
Interpretations     & 1215 (0.81)   & $0$ & 109 (0.76) & $0$ \\
Exploration         & 1010 (0.67)   & $0$ & 95 (0.66) & $0$ \\
\midrule
Valence (mean ± std) & 0.59 ± 0.34 &  & 0.49 ± 0.31 &  \\
Arousal (mean ± std) & 0.37 ± 0.19 &  & 0.30 ± 0.20 &  \\
\bottomrule
\end{tabular}%
}
\end{table}

It is worth noting that, for the \textit{who} cue in HRI scenarios, the second most frequent label was $1$ (corresponding to the ``you'' pronoun), with 65~(0.45) occurrences. Same for non-HRI with the second place for \textit{who} is label $2$ (another person) 643~(0.43). For non-HRI and emotional reactions, the second place is label \textit{weak} with 645~(0.43).

\subsection{Human-Robot Exchanges} 
\label{sec:human_robot_exchanges}
Finally, we demonstrate how \textit{whEE} can be used to modulate the empathy level of our social robot, Haru. We selected $308$ speaker utterances from the EDR and The Talking Room datasets and used them as input for Haru's behavior.

For each  utterance, Haru takes the listener role while its \textit{whEE}-based behavior determines whether the speaker is seeking empathy. Depending on this classification, we prompt Haru to generate either an empathetic or regular response. We then analyze the cues displayed in Haru's utterances.  

First, we applied our classification prompt with LLaMA~3.3 to determine whether each utterance was seeking empathy. For utterances classified as seeking empathy, we generated responses using our empathetic prompt; otherwise, we used Haru's regular prompt. This process resulted in $166$ empathetic responses, accounting for $53.9 \%$ of the total.

\begin{figure}
    \centering
    \includegraphics[width=1\linewidth]{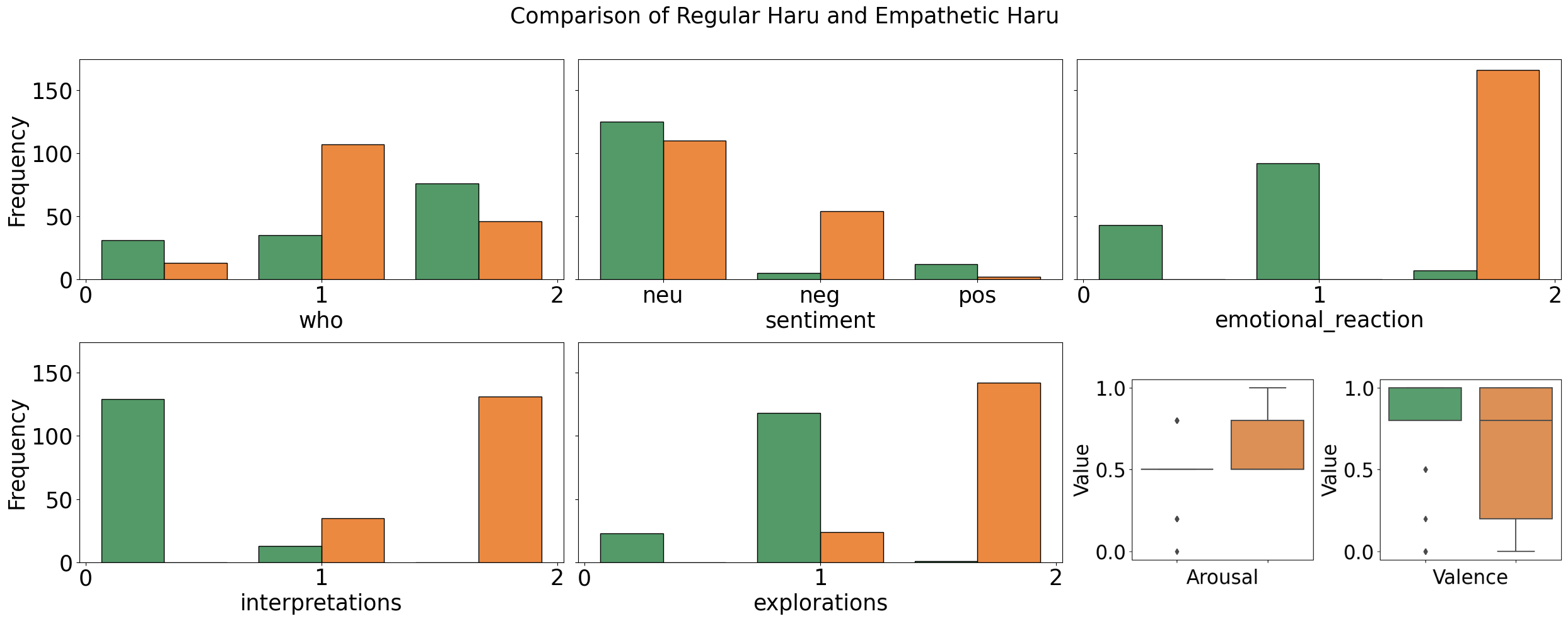}
    \caption{Empathy cues distribution comparing regular Haru (\textcolor{mygreen}{\textbf{green}}) and empathetic Haru (\textcolor{myorange}{\textbf{orange}}).}
    \label{fig:empathetic_haru}
\end{figure}

We then compared the empathy cues present in responses from regular and empathetic Haru using our classification prompt. The results of this comparison are shown in Figure~\ref{fig:empathetic_haru}.

Finally, we present a couple of examples of responses by Haru in EDR.

\begin{itemize}
    \item \textbf{Empathetic Haru}\\
    \textit{Speaker:}~[...] My girlfriend organized a surprise party for me. [...]~all my friends and family were there waiting for me for my party and I was it was so exciting.\\
    \textit{Haru:} I feel super happy for you~[...] I can imagine how thrilling it was to walk in and see all your loved ones waiting for you, what a wonderful girlfriend you have! Are you feeling still on cloud nine after that awesome celebration? 

    \item \textbf{Regular Haru}\\
    \textit{Speaker:} I'm about to go to my cousin's house~[...] I had no idea he could pay or afford a house in this neighborhood.\\
    \textit{Haru:} Wow, that's surprise! I'm curious, what's the house like? Is it super fancy?
\end{itemize}

In our repository, readers can find all the responses to the speaker utterances in EDR.

\section{Limitations}

The primary limitation of our framework is the scope of evaluation in diverse HRI scenarios and the lack of direct assessment of user experience, particularly comparing \textit{whEE} to Haru's current static behavior. Nonetheless, our results confirm that \textit{whEE} effectively identifies \textit{when} and \textit{how} robots should express empathy, indicating its potential for broader application. Besides, we focus on text-only empathy, which ignores other aspects of empathetic behavior. Future work will involve user studies in diverse HRI contexts and multimodal analysis.

\section{Discussion and Future Work} 
We demonstrate that LLMs can effectively identify when individuals are \textit{seeking} or \textit{providing} empathy, outperforming both classic AI and other deep learning methods in HRI. Deep learning methods like RoBERTa achieve a similar performance in HRI. Nevertheless, they perform worse than LLMs in the task of detecting \textit{seeking empathy}. Confusion matrices for all models are available in our repository.

LLaMA~3.3 showed strong accuracy in detecting \textit{seeking empathy} and \textit{none} labels ($0.81$ and $0.68$, respectively). However, accuracy drops to $0.51$ for \textit{providing empathy}, with $0.35$ of these utterances misclassified as \textit{none}. This suggests that detecting empathy provision may require more conversational context. Future work could explore using full dialogue exchanges as input to improve this classification, in a similar manner to~\cite{Montiel2024, Arzate2025_hri}. 

Utterances seeking empathy often show positive or negative sentiment, while empathetic responses tend to be neutral or positive. This indicates that people share varied experiences to seek empathy, but listeners avoid reinforcing negativity, aligning with findings on emotional modulation~\cite{Avenanti2006stimulus}.

In non-HRI empathy classification, transformer-based language models outperform LLMs and generalize well to HRI scenarios. However, LLMs underperform on our non-HRI test set, highlighting the need for further investigation.

Our empathetic generation scenario in HRI shows significant difference in empathy cues expressed by Haru. There are more attempts to refer to the empathy seeker, less neutrality in sentiment, and a wider emotional range. Likewise, we observe stronger expressions of explicit empathetic mechanisms. This, added to the fact that the robot uses this communication strategy only when empathy is being sought by a person, shows that using the whEE framework can be useful to avoid empathy burnout. Future work must be done to analyze how these responses are perceived and their effect. 

\section{Conclusions} 
We introduced \textit{whEE}, a framework enabling social robots to recognize \textit{when} empathy is needed and \textit{how} to respond empathetically. Leveraging LLMs and observable empathy cues, \textit{whEE} provides the HRI community with tools to develop robots that dynamically adjust empathy based on context.

Evaluation across diverse interaction scenarios—including dyadic and multi-group settings—demonstrated \textit{whEE}'s robust performance. \textit{whEE} consistently performs well in HRI scenarios using LLMs. Likewise, we tested its performance in generating empathetic responses when appropriate. These results validate the effectiveness of our design and highlight \textit{whEE}'s potential to support the development of socially intelligent robots with an enhanced socio-emotional understanding of human behavior.

\bibliographystyle{ieeetr}
\balance
\bibliography{Bibliography}

\end{document}